# RSV-SLAM: Toward Real-Time Semantic Visual SLAM in Indoor Dynamic Environments

Mobin Habibpour,[1,†] Alireza Nemati,[2,†,*], Ali Meghdari[1], Alireza Taheri[1], Shima Nazari[2,*]

[1] Social and Cognitive Robotics Lab., Sharif University of Technology, Tehran 1458889694, Iran.
[2] Department of Mechanical and Aerospace Engineering, University of California Davis, California 95618, USA
\* Corresponding authors: nemati@ucdavis.edu, snazari@ucdavis.edu
†Mobin Habibpour and Alireza Nemati contributed equally to this work

**Abstract.** Simultaneous Localization and Mapping (SLAM) plays an important role in many robotics fields, including social robots. Many of the available visual SLAM methods are based on the assumption of a static world and struggle in dynamic environments. In the current study, we introduce a real-time semantic RGBD SLAM approach designed specifically for dynamic environments. Our proposed system can effectively detect moving objects and maintain a static map to ensure robust camera tracking. The key innovation of our approach is the incorporation of deep learning-based semantic information into SLAM systems to mitigate the impact of dynamic objects. Additionally, we enhance the semantic segmentation process by integrating an Extended Kalman filter to identify dynamic objects that may be temporarily idle. We have also implemented a generative network to fill in the missing regions of input images belonging to dynamic objects. This highly modular framework has been implemented on the ROS platform and can achieve around 22 fps on a GTX1080. Benchmarking the developed pipeline on dynamic sequences from the TUM dataset suggests that the proposed approach delivers competitive localization error in comparison with the state-of-the-art methods, all while operating in near real-time. The source code is publicly available[1].

**Keywords:** Visual Simultaneous Localization and Mapping, Semantic Segmentation, Mobile Robot, Dynamic Environment

## Introduction

Simultaneous Localization and Mapping (SLAM) is a fundamental problem in mobile robotics, and it has become a vital prerequisite for robotic tasks. It consists of constructing a 3D map of the surrounding area while simultaneously estimating the robot's location and pose within the produced map. Many robotic applications - from navigation to manipulation - heavily rely on the SLAM algorithms to determine and visualize the robot's surrounding area[1]. Over the past few years, visual SLAM (vSLAM), has gained increasing attention due to the availability of high-end and cost-effective cameras and their richness of information in comparison to other onboard sensors[1]. Numerous notable vSLAM methods have been proposed using different types of input sensors, such as monocular SLAM [2] , RGBD SLAM[3], and stereo SLAM[4]. RGBD cameras are popular in indoor environments due to their direct depth map and metric scale.

---
[1] https://github.com/mobiiin/rsv_slam

The current vSLAM methods have demonstrated success in some scenarios, but most of these systems assume the surrounding world to be static. Estimation of the pose reconstruction of the map are negatively affected by dynamic and moving objects, such as animals, people, and vehicles. Therefore, rendering the standard vSLAM systems unsuitable for real-world scenarios[1]. It is important to note that traditional vSLAM systems are still capable of achieving satisfactory performance when dynamic elements are in the minority[5]. Estimation methods such as RANSAC or robust cost functions are used to overcome that problem by classifying the dynamic features as outliers[6]. Nevertheless, these modules only provide limited improvement, because they are limited to handling static scenes with a slight degree of dynamic, and they may still fail when dynamic objects occlude the view. As an instance, when the moving objects account for a major proportion of the features in a scene, since these dynamic features are not in the minority, they cannot be excluded as outliers. Due to these dynamic data, the SLAM process will either suffer from drift and inaccurate estimated trajectory or even lead to failure. In addition, these dynamic objects corrupt the structured 3D map. The adverse effect of dynamic objects on the 3D reconstructed map is shown in Fig. 1. Consequently, vSLAM systems need to be upgraded to accommodate highly dynamic environments[1].

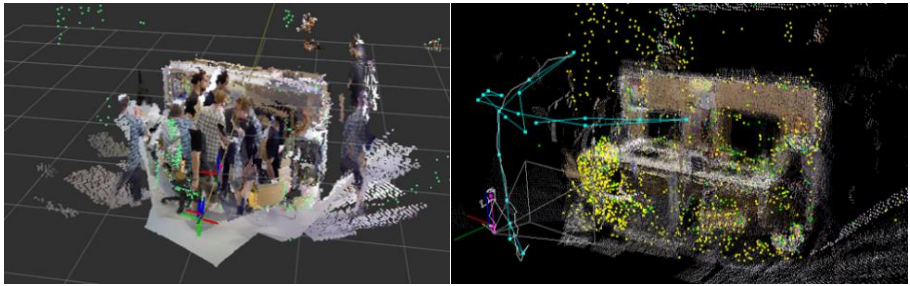

**Fig. 1.** Corrupted point cloud due to dynamic objects in the scene in the Rviz environment on the left(TUM sequences [7] fr3/walking static). The point cloud free of dynamic objects, created on the same dataset using the RSV-SLAM on the right.

A few vSLAM methods have been proposed to handle dynamic objects in the environment. All these methods either have high computational needs, which makes them unsuitable for real world robotic applications, or they require a disproportionally high level of accuracy and processing power. In this paper, we present a lightweight modular Real-time Semantic Visual SLAM (RSV-SLAM) system capable of detecting and tracking dynamic objects and image inpainting while keeping the final frame-per-second near real-time. ROS ensures the modularity of the framework, which is essential for robotic implementation. In addition to facilitating the future replacement and coupling of different components to improve performance, this feature also allows for run-time management in case an issue arises[8][9][10].

In this paper, we make the following principal contributions:
- A near real-time semantic RGBD SLAM system, which can reduce the influence of moving objects in dynamic environments.
- Integration of a tracking module for estimating the velocity of dynamic objects and utilizing their feature points in cases where they are temporarily static.
- A generative algorithm integration to inpaint the missing sections of input frame before 3D reconstruction.
- ROS integration to facilitate the robotic implementation process.

The remaining portion of the document is organized as follows. Section II presents a review of previous research in the field. Section III provides an overview of the system and outlines the details of each unique module. Section IV shows experimental results, and section V presents the conclusions and discusses future work.

## Related Work

In recent years, SLAM problems in dynamic environments have gained attention [1]. Approaches developed to mitigate this problem usually classify the features into several groups. The feature points, which are assured to be static, are used for the SLAM process. These approaches can be roughly classified into two types: Geometry-based and Semantic-based solutions[11].

Geometry-based methods reject dynamic objects using either weight functions or motion consistency constraints, by classifying the dynamic features as outliers. Sun et al. [12] proposed Particle Filtering as a method of removing motion patches from images. Kim et al. [13] modeled the background nonparametrically, using this model to estimate the odometry. StaticFusion [14] constructs the background by using probabilistic segmentation and weighted dense optimization on the image. Flowfusion [15] detects dynamic regions in the scene using dense optical flow residuals and reconstructs the static world using a an identical method as StaticFusion.

Geometric methods do not require a prior knowledge of moving objects and require relatively low computational power. They are based on the geometrical movements of features and have no regard for the semantic information of the scene. However, they are unable to detect dynamic objects that are temporarily static, and they are generally less accurate compared to semantic-based approaches. The geometry-based approaches are also incapable of building semantic maps of the environment[11].

Using deep neural networks, learning-based methods are capable of detecting pixel-wise labels andproviding masks from input images to extract semantic information from the environment. Semantic masks can then be used to identify potentially dynamic objects in a single image without the need to process multiple frames. Some approaches try to combine geometry constraints with semantic information to improve the outcomes. DS-SLAM[6], built on top of ORB-SLAM2 [3], incorporates a semantic segmentation network (SegNet)[16] as well as moving consistency checks to reduce the impact of dynamic objects.

To address a wide range of objects, DynaSLAM [17] utilizes Mask-RCNN [18] augmented with multi-view geometry. Through deep learning, multi-view geometry, or both, it can detect moving objects and inpaint the background when it has been obscured by moving objects. Although this method can provide good results on publicly available datasets, its computation cost is relatively high; therefore, it is primarily used offline. The multi-object tracking capability of DynaSLAM II [19] is tightly integrated. However, this method just works for static objects. Nevertheless, in the dynamic and moving scene of the TUM [7] dataset, people outline constantly change shape as they walk and sit.

Instead of simply removing dynamic objects, some methods continuously track their motion. MID-fusion [20] proposes a tracking algorithm to track both object poses and camera motion. In contrast, [21] presents a new motion model for tracking Inflexible moving objects without prior knowledge of their 3D model. Despite their significant improvements in dynamic environments, these approaches are neither suitable for robotic implementation nor running in real-time. Our approach, on the other hand, prioritizes real-time performance and ROS-enabled frameworks to facilitate robotic implementation.

## System Overview

We employed a combination of semantic and geometry information extracted from RGBD input images to address the challenges posed by moving objects in dynamic environments. In particular, our approach utilizes learning-based methods to process potentially dynamic objects and incorporates an Extended Kalman Filter module to prevent removing dynamic objects that are temporarily idle. To ensure real-time performance, which is essential for robotic applications, we implemented a state-of-the-art instance segmentation framework optimized for real-time performance. The proposed system has been built on RTAB-Map [5], a feature-based SLAM system primarily designed for static environments. Furthermore, an image inpainting module is used to fill in the areas removed by the segmentation module. The overall framework is illustrated in Fig. 2, with detailed explanations of the semantic, tracking, and filling modules provided in the subsequent sections.

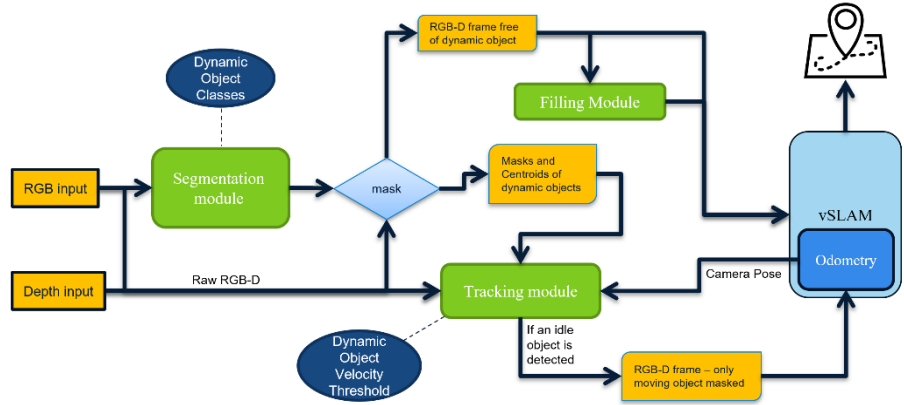

**Fig. 2.** The overall framework of the RSV-SLAM method. The stages we have added to the SLAM system are represented by boxes shaded in green. These additional stages consist of a segmentation module, a tracking module, and a filling module.

### 3.1 Segmentation Module

Using deep learning-based methods, instance segmentation generates pixel-level labels and masks of identified objects in the RGB images. A number of open source, state-of-the-art instance segmentation frameworks were selected to implement and see their performance in action[22][23][16][18]. Considering both accuracy and computational speed, we opted for the recently improved YolactEdge [22] network, which is an instance segmentation framework. In this network, a fully connected network provides the masks, while the convolutional network is responsible for computing the corresponding coefficients along with the object detection task. This technique allows for higher computational speed with relatively competitive accuracy. The ResNet-50-FPN backbone is utilized for the YolactEdge architecture since it delivers better real-time performance[22].

Using more advanced deep neural networks such as Mask-RCNN [18] can lead to more accurate segmentation outcomes, but it comes at the expense of increased computational costs. The segmentation model utilized in this study was pre-trained on the Microsoft COCO dataset [24], which includes a diverse range of 91 object classes. We selected our objects of interest based on the dynamic objects present in TUM sequences and the deployment environment of our robot. Therefore, we chose highly dynamic object classes that are common in indoor environments for segmentation such as human, chair, bottle, and so on.

The segmentation module receives an RGB image as well as produced bounding boxes, classes, and confidence scores and binary masks per instance. The segmentation module also generates centroids for each detected dynamic object using the bounding box coordinates and the depth information; the tracking module utilizes these centroids and their 3D coordinates for velocity estimation.

### 3.2 Tracking Module

The tracking module employs the bounding box data from the segmentation module and the odometry information from RTAB-Map to predict the velocity of dynamic objects. We adopted the EKF algorithm introduced by Vincent et al [25]. First, the tracking module receives the segmentation masks as well as the object detection data, including bounding box coordinates. Using the center of each bounding box and the corresponding depth pixel, centroids containing the 3D world coordinates for each dynamic object are calculated. Essentially, the 3D centroid of an object is located at the center of its bounding box. By seeing the tracked object again, its 3D centroid coordinates are updated; otherwise, the predicted position is used. We do not update the mask of dynamic objects based on the predicted positions.

Second, an EKF set of matrices is instantiated for each new dynamic object (based on type of the object, the binary mask of the object as well as its position). The objects are constantly tracked, and their hidden state vectors are updated using the EKF algorithm.

Third, an object is categorized as moving if the calculated velocity exceeds the class threshold, otherwise, it is classified as temporary static. Therefore, the corresponding mask is updated accordingly. The mask of the objects, which are decided to be idle are removed to make use of their feature points for pose estimation and camera tracking. This is beneficial when the camera is moving in the same direction and at the same speed as other objects. A good example would be a moving car on the highway. Another useful scenario would be camera occlusion by an idle object with a high number of features.

It is worth noting that the RGBD images containing idle objects are only fed to the RTAB-Map's odometry module; therefore, feature points belonging to dynamic objects will not be used for map construction. Objects that are not observed for a few frames are considered departed from the environment, their filter is removed, and their tracking is terminated. This module additionally calculates the intersection over union (IoU) of the dynamic objects in the consecutive frames. If the object is non-static having an IoU of higher than a certain threshold, it will be considered dynamic, regardless of its velocity.

### 3.3 Filling Module

The filling module receives the segmented images free of dynamic objects from the segmentation module and proceeds to inpaint the removed areas of the picture. A number of open source, state-of-the-art image inpainting frameworks were selected to implement and see their performance in action[26][27]. After comparing their performance and results, we adopted the CR-Fill framework [27], which uses a generative adversarial network for image inpainting. This work introduced an auxiliary contextual reconstruction branch/loss (CR loss) to the generator network to choose more appropriate image patches for the missing regions. We employed the model, pretrained on the Places2 training set [28], provided by Zeng et al. [27].

# Experimental Results

The TUM RGBD dataset[7], which is widely recognized in the Simultaneous Localization and Mapping (SLAM) community for evaluating RGBD SLAM, was used to test the proposed method. We compared the obtained results with the state-of-the-art dynamic vSLAM methods. Additionally, a runtime analysis is provided to illustrate the efficacy of RSV-SLAM method. Finally, we demonstrated the effectiveness of RSV-SLAM on a real robot under real world dynamic scenarios.

All experiments are conducted on a computer equipped with GTX1080 GPU. We tested the RSV-SLAM algorithm on Ubuntu 18.04 using the Robot Operating System Melodic (ROS). All the modules are implemented as separate components on the ROS platform.

**Table 1.** Experimental parameters used for segmentation and tracking modules.

| Description | Amount |
|---|---|
| Frame termination for object tracking | 10 |
| Threshold score | 0.9 |
| Maximum tracked objects | 5 |
| Threshold velocity for people | $0.7 m/s^2$ |
| Threshold velocity for objects | $1.2 m/s^2$ |

Table 1 presents the parameters employed in the segmentation and tracking modules. These parameters are estimated empirically, based on evaluating TUM sequences and detecting real-world objects. By setting a score threshold and limiting the number of tracked objects, the robustness of the performance is ensured.

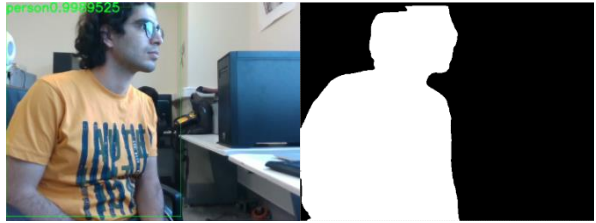

**Fig. 3.** Original RGB image including the object detection on the left. The calculated binary mask on the right.

Fig. 3 shows the mask output of the segmentation module. The unsatisfactory performance of the instance segmentation framework in occluded scenes is its biggest limitation. Fig. 4b illustrates the output of the inpainting module on the TUM dataset. While there are some visible artifacts surrounding the desk and chair, the network delivers satisfactory inpainting results. The inpainting module will perform well as long as dynamic objects are in the minority.

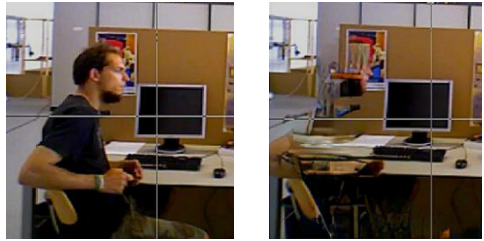

(a)　　　　　(b)

**Fig. 4.** Qualitative results of the Inpainting module on TUM dataset[7]. (a) before inpainting, (b) after inpainting using [27]

### 4.1 Evaluating Performance Against Current Best Practices

We compared the purposed approach with the current state-of-the-art learning-based methods, such as MID-Fusion [20], DS-SLAM [6], and DynaSLAM [17], and the results are summarized in Table 2. Although RSV-SLAM does not outperform DynaSLAM, it still delivers competitive results in dynamic sequences. However, DynaSLAM's Mask-RCNN network and it's region-growing algorithm are computationally intensive, making it impractical for real-time applications due to the time-consuming computations involved. Whereas we are able to reach near real-time operations and still provide very close results to the previous methods.

As shown in Table 2, the proposed method achieves similar results to RTAB-Map in slightly more dynamic scenarios. Whereas RSV-SLAM provides more accurate results when dealing with highly dynamic sequences. As demonstrated in Fig. 5d, RTAB-Map either has substantial drift or even fails in some situations. By removing the dynamic parts, we can estimate the camera pose more accurately, as illustrated in the trajectory estimation results in Fig. 5a through c.

**Table 2.** The RMSE of Absolute Trajectory Error (ATE) was compared, with the best results being highlighted in bold and the second-best results being underlined. When available, the results published in the original papers were utilized for the comparison.

| TUM Seqs | Notable Approaches | | | | |
|---|---|---|---|---|---|
| | RTAB-Map | MID-Fusion | DS-SLAM | DynaSLAM | RSV-SLAM |
| fr3/sit_static | 0.008 | 0.01 | <u>0.0065</u> | - | **0.003** |
| fr3/sit_xyz | 0.02 | 0.062 | - | **0.015** | <u>0.018</u> |
| fr3/wlk_static | 0.283 | 0.023 | <u>0.0081</u> | **0.006** | 0.063 |
| fr3/wlk_xyz | 0.442 | 0.068 | <u>0.0247</u> | **0.015** | 0.055 |
| fr3/wlk_halfs. | 0.164 | 0.038 | <u>0.0303</u> | **0.025** | 0.068 |
| fr3/wlk_rpy | 0.29 | - | 0.4442 | **0.035** | <u>0.101</u> |

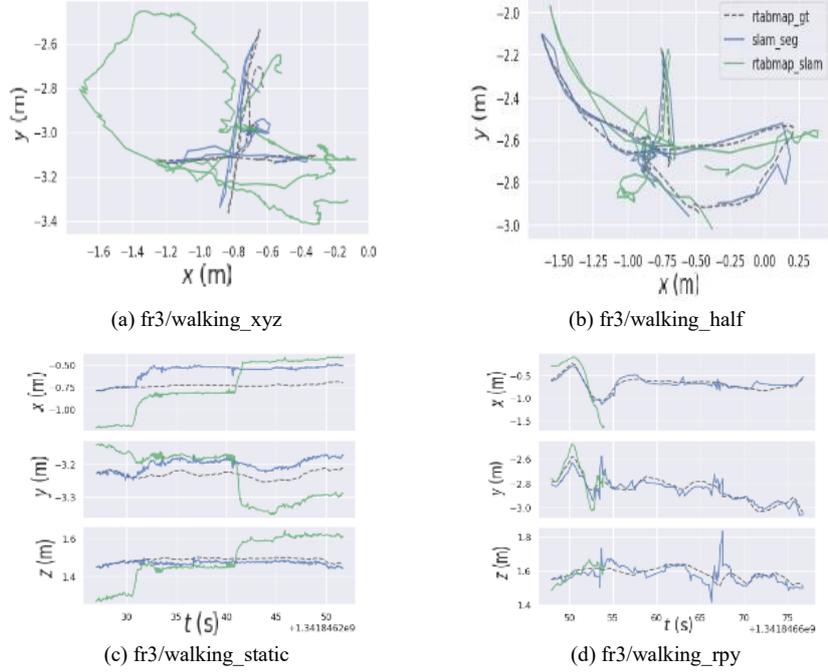

(a) fr3/walking_xyz  (b) fr3/walking_half

(c) fr3/walking_static  (d) fr3/walking_rpy

**Fig. 5.** Estimated trajectories by RTAB-Map (green line) and the RSV-SLAM (blue line) against ground truth (grey dotted line).

**Table 3.** Comparison of Computation Time [ms]

| Approach | Average Time | GPU |
|----------|--------------|-----|
| RTAB-Map | 35 | - |
| DS-SLAM | 148.53 | - |
| DynaSLAM | 1144.93 | - |
| RSV-SLAM | 60 | GTX1080 |

### 4.2 Runtime Analysis

We evaluated the proposed method by measuring its average computation time and comparing it to the state-of-the-art and learning-based approaches such as DS-SLAM [6] and DynaSLAM [17], as well as the baseline RTAB-Map[5]. Regarding the results presented in Table 3, it can be concluded that the proposed method stands out as the sole semantic RGBD SLAM solution capable of achieving near real-time performance on a relatively old GPU system. DynaSLAM not only relies on Mask-RCNN[18], which is a two-stage segmentation framework, but it also suffers from region growing algorithm which are time-consuming and can significantly delay frame processing when dealing with numerous dynamic features. DS-SLAM uses the segnet[16] network for the segmentation task, and it only deals with moving people. All the mentioned frameworks are working on 640×480 image resolution.

### 4.3 Real World Implementation

We further tested the proposed method employing the Arash 2.0 social robot equipped with Intel RealSense(D435) RGBD from to assess its effectiveness in the real-world dynamic environment. Fig. 6a illustrates the Arash 2.0 social robot, designed and fabricated by the social cognitive robotic team at Sharif University of Technology. In the conducted experiments, a person was constantly sitting and walking in front of the robot while the robot was moving around. In order to provide a qualitative basis for comparison, we first ran the RTAB-Map SLAM in a static environment, then we evaluated both the RTAB-Map and the developed framework in the same environment in the presence of a moving person.

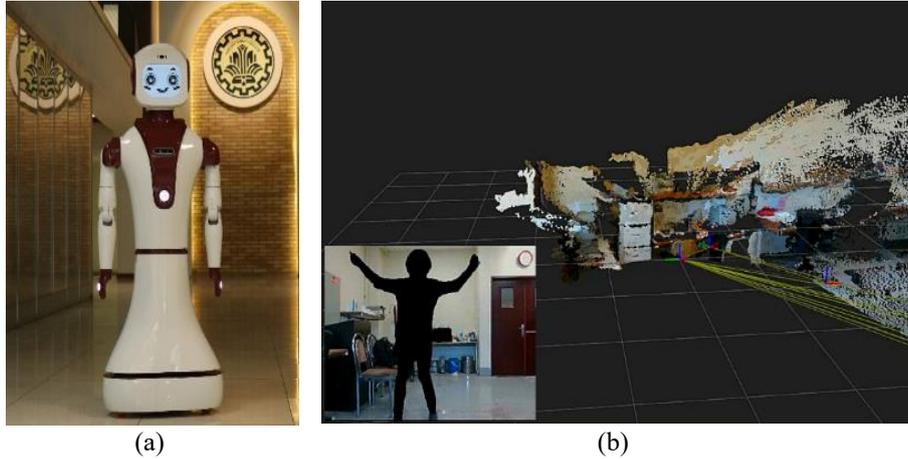

(a)                  (b)

**Fig. 6.** (a) The Arash 2.0 Social Robot. (b) The Rviz software environment, showing the constructed map of the robotic lab

Fig. 6b shows the ROS visualization software called Rviz. The inset picture is the input image, free of dynamic objects. As shown in Fig. 6b, dynamic objects have been filtered out using the masks created by the segmentation and tracking modules.

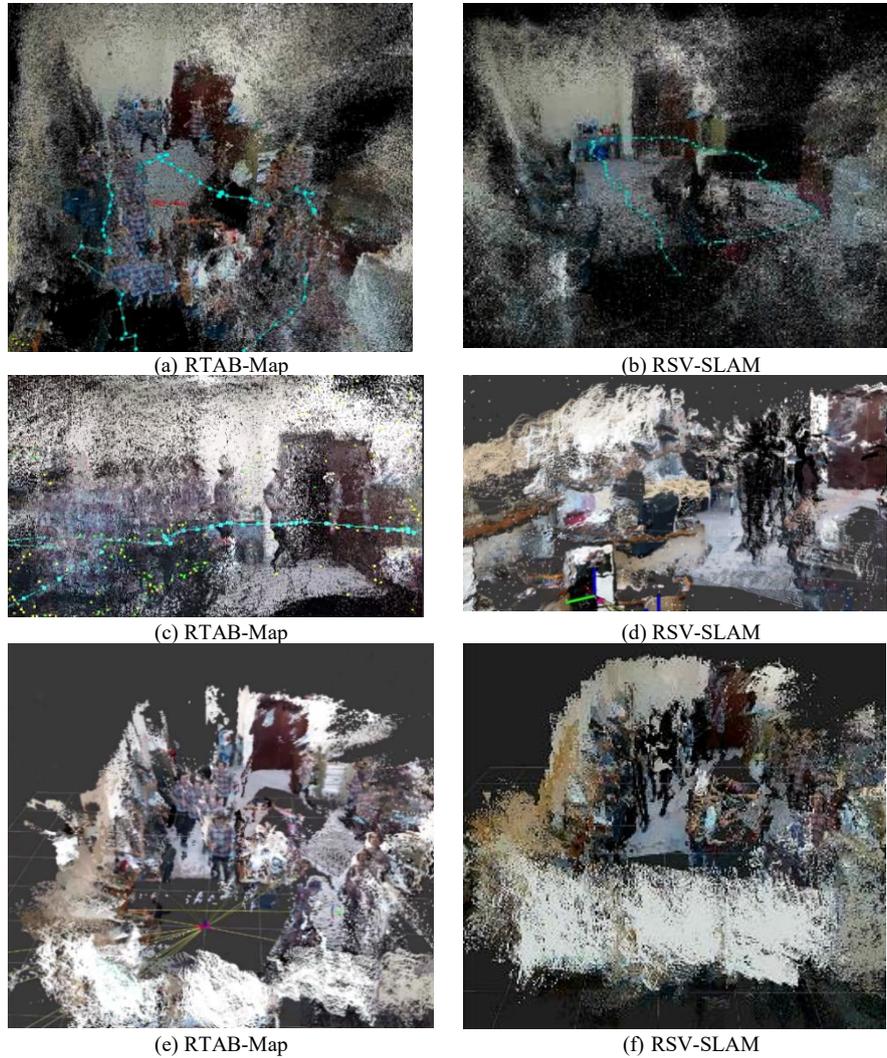

**Fig. 7.** Evaluation of SLAM frameworks in the Sharif CEDRA Social Cognitive Robotics lab. The left column shows the constructed map and odometry by RTAB-Map in dynamic scenario. The outcome of the our proposed method in the same scene is demonstrated in the right column.

Fig. 7 presents the qualitative comparison of the proposed method against the RTAB-Map. Fig. 7a illustrates the devastating effects of dynamic objects on the odometry calculation. The robot in this scenario had a smooth trajectory around the lab. Through removing and ignoring the feature points belonging to the dynamic objects, the RSV-SLAM effectively calculates the odometry of the robot, as shown in Fig. 7b. Fig. 7c shows the adverse effect of a moving person on the constructed 3D map. RTAB-Map fails to capture the tables on the left wall of the lab due to the presence of a dynamic agent in the scene. However, Fig. 7d demonstrates the effectiveness of our approach in minimizing the detrimental effects of dynamic agents on the reconstructed map.

Dynamic agents may cause the SLAM process to incorrectly close a loop, or even prevent a correct loop from being detected. Fig. 7e shows that, the robot has reached a previously seen location but has failed to detect and close the loop. Consequently, the accumulated drift is presented in the incorrectly constructed wall. Fig. 7f, conversely, shows the RSV-SLAM successful in performing loop closure and capturing the rectangular shape of the lab. It can be concluded that the proposed framework successfully constructs the static map of the laboratory while maintaining accurate odometry.

## Conclusion

In this study, we presented the RSV-SLAM, a real-time semantic RGBD SLAM framework suitable for implementation in real-world robotics applications. The segmentation module is augmented by an Extended Kalman Filter module to remove the masks off temporary idle objects. A generative image inpainting module complements the framework to fill in the unknown areas of the input image. We developed a fully modular, interchangeable, and implementation-ready framework on the ROS platform that achieves around 22 fps on a GTX1080 GPU. An extensive evaluation of RSV-SLAM demonstrates its ability to provide competitive localization accuracy while still operating in near real-time. The TUM dataset assessment of the proposed methodology reveals performance that is on par with state-of-the-art methods. Even though RSV-SLAM does not outperform DynaSLAM on localization accuracy, it offers better real-time performance with a semi-dense 3D map that is devoid of dynamic objects. In future works, we hope to improve the instance segmentation performance when dealing with occluded scenes. Fusion of the 3D LIDAR data with the RGB-D inputs could assist us in constructing more accurate static maps of dynamic environments.